\documentclass{article}

\usepackage{arxiv}
\usepackage[utf8]{inputenc}
\usepackage[T1]{fontenc}
\usepackage{hyperref}
\usepackage{url}
\usepackage{booktabs}
\usepackage{amsfonts}
\usepackage{amsmath}
\usepackage{amssymb}
\usepackage{mathtools}
\usepackage{amsthm}
\usepackage{nicefrac}
\usepackage{microtype}
\usepackage{graphicx}
\usepackage{subcaption}
\usepackage{algorithm}
\usepackage{algorithmic}
\usepackage{pifont}
\usepackage{fancyvrb}
\usepackage{fvextra}
\usepackage[numbers]{natbib}
\usepackage{doi}
\usepackage{enumitem}
\usepackage{xcolor}
\usepackage{textcomp}

\theoremstyle{plain}

\theoremstyle{definition}

\theoremstyle{remark}

\title{ReTreVal: Reasoning Tree with Validation and Cross-Problem Memory for Large Language Models}

\author{
  Abhishek HS \\
  QpiAI, Bengaluru, India \\
  \texttt{abhishek.hs@qpiai.tech}
  \And
  Pavan C Shekar \\
  QpiAI, Bengaluru, India \\
  \texttt{pavan.s@qpiai.tech}
  \And
  Arpit Jain \\
  QpiAI, Bengaluru, India \\
  \texttt{arpit.j@qpiai.tech}
  \And
  Aswanth Krishnan \\
  QpiAI, Bengaluru, India \\
  \texttt{ashwanth.krishnan@qpiai.tech}
}

\date{}

\begin{document}
\maketitle

\begin{abstract}
Every existing inference-time reasoning framework discards all failure context at problem
boundaries, leaving a model solving problem 500 no wiser than it was on problem 1.
We present \textbf{ReTreVal} (Reasoning Tree with Validation), a training-free framework
that closes this gap through adaptive tree exploration with tool-augmented node refinement,
typed-failure backtracking that injects categorized error context into the recovered branch,
and a self-rewriting memory that accumulates and revises strategy entries across
problems, enabling inference-time cross-problem learning on any fixed, unmodified LLM
without fine-tuning.
ReTreVal achieves \textbf{85.8\%} pass@1 on MATH-500
(+8.6\,pp over Zero-Shot CoT, +8.6\,pp over the strongest baseline Self-Refine)
and \textbf{54.4\%} on MMLU-Pro (+15.3\,pp over Self-Refine),
with a \textbf{3.4:1} win-to-regression ratio confirming genuine error recovery rather
than noise. These capabilities, previously requiring gradient updates, allow a 32B model
to compete with much larger single-pass systems.

\textbf{Code:} \url{https://github.com/qpiai/retreval}
\end{abstract}


\section{Introduction}

When a model fails to solve a problem, the failure is informative, revealing which
approach was tried, where reasoning broke down, and what class of error was made.
Yet every existing inference-time reasoning framework discards this information the
moment a problem ends.
A model attempting problem 500 of a benchmark begins from precisely the same
uninformed state as it did on problem 1, despite having encountered, and potentially
recovered from, hundreds of structurally similar failures in between.
This raises a concrete research question: \emph{can a fixed, unmodified LLM accumulate
typed failure experience across a problem sequence and act on it, without any parameter
update?}

Prior work, including ReAct~\citep{yao2023react}, Reflexion~\citep{shinn2023reflexion},
Self-Refine~\citep{madaan2023self}, Tree-of-Thoughts~\citep{yao2023tree}, and
LATS~\citep{zhou2023language}, addresses individual failure modes \emph{within} a single
problem but discards all failure context at problem boundaries and retries blindly without
categorising why a branch failed.
The gap is architectural. No existing framework provides the typed failure accumulation,
cross-problem memory, or inference-time self-improvement needed to answer the question above.

We introduce \textbf{ReTreVal} (Reasoning Tree with Validation), a training-free framework
that answers this question affirmatively.
Five tightly coupled mechanisms, namely adaptive tree construction, adaptive tool-augmented
refinement, typed-failure backtracking, critique-driven dual scoring with skepticism
penalties, and self-rewriting memory, are described in detail in
Section~\ref{sec:architecture}.
Together they enable inference-time cross-problem learning, so a model running ReTreVal on
problem 500 has genuinely refined its strategies based on what failed on problems 1--499,
without any weight update.

We validate on MATH-500~\citep{hendrycks2021measuring} and
MMLU-Pro~\citep{wang2024mmlu}.
ReTreVal achieves \textbf{85.8\%} on MATH-500 (+8.6\,pp over Self-Refine) and
\textbf{54.4\%} on MMLU-Pro (+15.3\,pp), with a \textbf{3.4:1} win-to-regression ratio
confirming genuine error recovery rather than noise.

\paragraph{Contributions.}
\begin{itemize}[leftmargin=1.5em, itemsep=2pt]
  \item \textbf{Inference-time cross-problem learning.}
        A demonstration that a fixed, unmodified LLM can improve its reasoning
        accuracy across a problem sequence purely at inference time, via self-rewriting
        memory with typed failure accumulation, LLM-driven entry revision, and
        confidence-weighted retrieval, with no fine-tuning or reward model.
  \item \textbf{Typed-failure backtracking.}
        Failures are categorised by root cause and approach type, then injected into the
        sibling branch's context, structurally preventing repetition of the same error class.
  \item \textbf{Tool-augmented adaptive refinement.}
        An extensible tool registry (symbolic solver, arithmetic evaluator, combinatorics
        engine, LP solver) is dispatched per-node inside the tree, eliminating the
        computation errors that dominate baseline failures.
  \item \textbf{Empirical validation.}
        Consistent gains over all baselines on MATH-500 and MMLU-Pro, with a 3.4:1
        win-to-regression ratio confirming genuine error recovery.
\end{itemize}

\section{Related Work}
\label{sec:related_work}

\paragraph{Iterative Reasoning.}
ReAct~\citep{yao2023react} interleaves thought-action-observation cycles but its linear
structure limits alternative path exploration and retains no cross-problem memory.
Reflexion~\citep{shinn2023reflexion} adds verbal self-reflection after each failed attempt,
storing insights in short-term episode memory, but its trial-and-error search lacks
structured exploration.
Self-Refine~\citep{madaan2023self} iteratively critiques and rewrites a single solution
but operates on one path with no persistent memory across problems.
LLM-as-judge paradigms~\citep{zheng2023judging} and
self-consistency~\citep{wang2023self} improve answer selection but require trained critics
or large sample budgets.

\paragraph{Tree Search Methods.}
Tree-of-Thoughts (ToT)~\citep{yao2023tree} generates multiple candidate thoughts and
selectively expands promising paths, but lacks node-level refinement and cross-problem
memory.
LATS~\citep{zhou2023language}, RAP~\citep{hao2023reasoning}, and
ReAcTree~\citep{choi2025reactree} combine tree search with self-reflection, Monte Carlo
planning, or hierarchical subgoal expansion, but none integrate persistent memory nor
continuous tool-augmented refinement at each node.

\paragraph{Memory and Learning.}
Voyager~\citep{wang2023voyager}, Generative Agents~\citep{park2023generative},
MemoryBank~\citep{zhong2023memorybank}, and MemGPT~\citep{packer2023memgpt} maintain
cross-episode memory streams for embodied agents, but do not transfer to pure reasoning
tasks.
TextGrad~\citep{yuksekgonul2024textgrad} propagates textual gradients through a fixed
pipeline to optimize prompts; ReTreVal pursues a complementary goal of evolving a dynamic
knowledge base across problems rather than optimizing a fixed pipeline, entirely at
inference time.

\paragraph{Positioning.}
Two capabilities in ReTreVal have no direct precedent.
\textbf{Typed-failure backtracking}: failures are categorised by root cause and approach
type, then injected into the sibling branch so the same error class cannot recur; all
prior tree-search methods retry with generic reflection.
\textbf{Inference-time cross-problem learning}: the self-rewriting memory accumulates
experience signals, triggers LLM-driven self-rewriting after repeated failures, and
tracks per-approach-type reliability; no prior framework enables a fixed model to improve
its strategy across a benchmark run without parameter updates.

\section{ReTreVal Architecture and Methodology}
\label{sec:architecture}

\subsection{System Overview}

\begin{figure}[!t]
  \centering
  \includegraphics[width=0.85\linewidth]{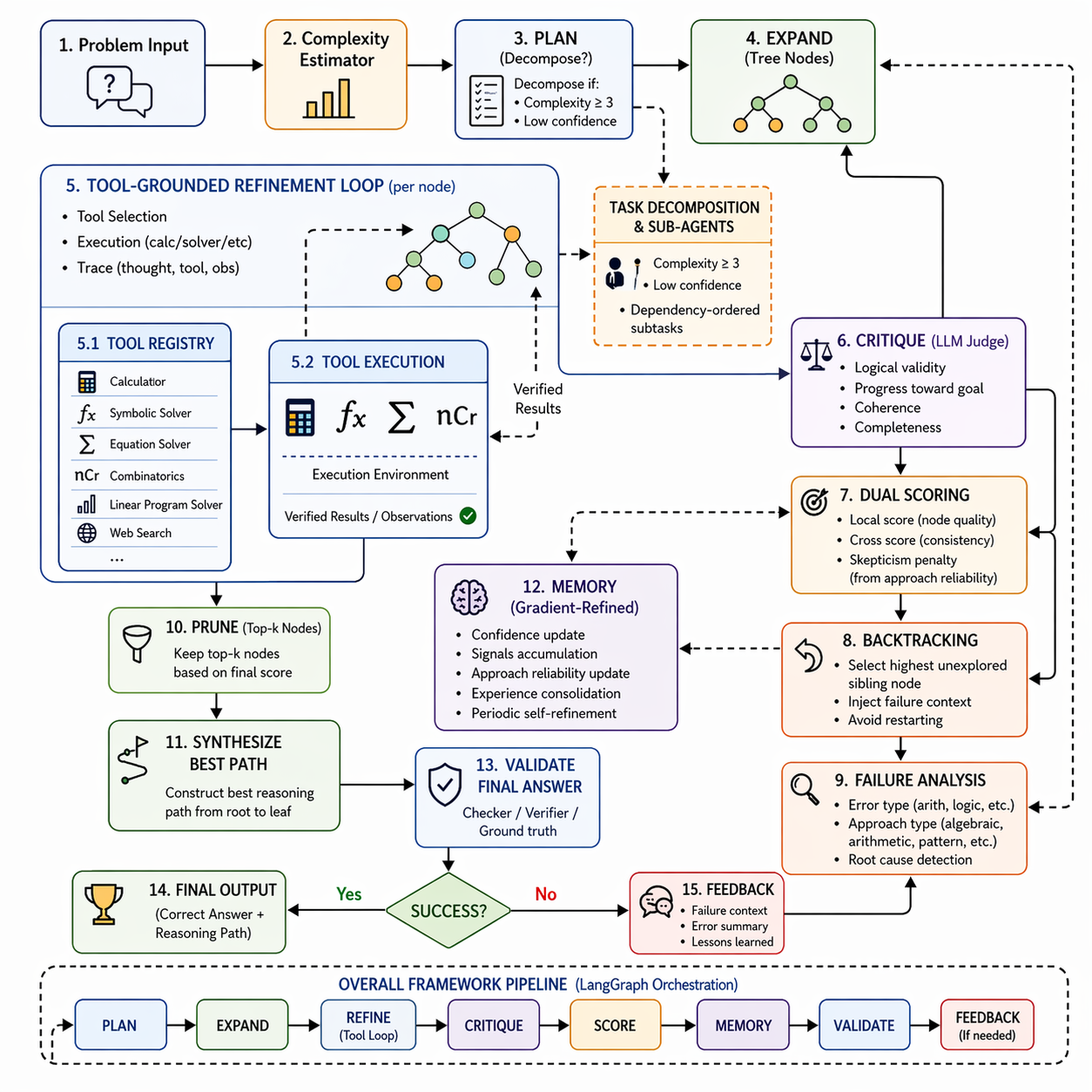}
  \caption{ReTreVal architecture overview. The framework combines adaptive tree-based
           reasoning with adaptive tool-augmented refinement at each node, LLM-based
           dual critique scoring with skepticism penalties, intelligent backtracking with
           failure analysis, self-rewriting memory, and task decomposition for
           high-complexity problems. Solid arrows indicate the main pipeline flow;
           dashed arrows indicate the backtracking recovery path triggered on validation
           failure.}
  \label{fig:architecture}
\end{figure}

ReTreVal is orchestrated as a LangGraph~\citep{langgraph2024} state graph
(Figure~\ref{fig:architecture}) with a fixed pipeline:
\textsc{Plan} $\rightarrow$ \textsc{Expand} $\rightarrow$ \textsc{Refine} $\rightarrow$
\textsc{Critique} $\rightarrow$ \textsc{Score} $\rightarrow$ \textsc{Memory} $\rightarrow$
(iterate or synthesize) $\rightarrow$ \textsc{Validate} $\rightarrow$ \textsc{Feedback},
with conditional edges for backtracking and decomposition.
The six components below are deliberately coupled. Failure context produced by
Backtracking flows directly into sibling Critique histories; memory insights are injected
into every Expand prompt; and the skepticism penalty in Scoring is updated by the same
memory that tracks per-approach-type reliability.

\subsection{Adaptive Tree Construction}

Fixed-depth tree methods (e.g., ToT at depth 2) either under-explore hard problems or
waste compute on easy ones.
ReTreVal addresses this by having the LLM estimate problem complexity on a 1--5 scale
\emph{before} tree construction; depth (2--5) and branching factor (2--4) are then set
from Table~\ref{tab:tree_params}. This single planning-time call adds negligible overhead
relative to tree expansion.

\begin{table}[ht]
\centering
\caption{Adaptive tree parameters by problem complexity.}
\label{tab:tree_params}
\renewcommand{\arraystretch}{1.15}
\begin{tabular}{|l|c|c|}
\hline
\textbf{Complexity} & \textbf{Max Depth} & \textbf{Children/Node} \\
\hline
1--2 (Simple)  & 2    & 2 \\
3 (Moderate)   & 3    & 3 \\
4--5 (Complex) & 4--5 & 4 \\
\hline
\end{tabular}
\end{table}

At each iteration, leaf nodes below max depth generate 2--4 child thoughts guided by
memory insights and prior failure patterns. After scoring, the top-$k$ nodes are retained
and lower-scoring branches pruned according to:
\[
  \mathcal{L}_{t+1}
  = \operatorname{top}_k\!\Bigl(
      \bigl\{n \in \mathcal{L}_t : \mathrm{depth}(n) < d_{\max}\bigr\},\;
      \mathrm{score}(n)
    \Bigr)
\]
where $\mathcal{L}_t$ is the frontier at iteration $t$. This best-first strategy
concentrates refinement compute on the most promising paths rather than expanding all
leaves uniformly.

\subsection{Adaptive Tool-Augmented Refinement}

Pure language model reasoning has no mechanism to verify intermediate computations. An
incorrect numerical result produced mid-chain is indistinguishable from a correct one,
and subsequent steps build on it silently.
ReTreVal addresses this through an adaptive tool-augmented refinement loop
(Algorithm~\ref{alg:react_refine}): at each step the LLM dynamically selects whichever
tool is most appropriate for the current sub-problem (symbolic equation solver, arithmetic
evaluator, combinatorics engine, or LP solver), executes it, observes the verified result,
and continues until it emits \texttt{FINISH} or a safety ceiling is reached.
This parallels the tool-augmented-reasoning lineage of
Toolformer~\citep{schick2023toolformer} and PAL~\citep{gao2023pal}, which similarly
offload computation to external tools; ReTreVal differs in that tool use is adaptive
per-node inside a tree, rather than a linear chain.

Crucially, the tool registry is fully extensible. Domain-specific tools (e.g., a chemistry
equation balancer, a geometry calculator, or a custom knowledge lookup) can be registered
as plain callables without modifying any part of the framework. The LLM only invokes the
tool it judges relevant to the current step; irrelevant tools in the registry incur no
cost. This separation of reasoning (LLM) from computation (tools) lets the system scale
its problem-solving capability simply by expanding the tool set, with no retraining or
architectural changes required.

Three guards prevent runaway execution via exact-duplicate call detection, a
consecutive-error threshold of 3, and an absolute ceiling of $S+2$ LLM calls.

\begin{algorithm}[ht]
\caption{Tool-Augmented Node Refinement}
\label{alg:react_refine}
\begin{algorithmic}[1]
\STATE \textbf{Input:} node $n$, problem $P$, plan, memory summary $M$,
       tool registry $\mathcal{T}$, max steps $S$
\STATE \textbf{Output:} refined node $n'$, execution trace $\tau$
\STATE $\tau \leftarrow [\,]$, calls $\leftarrow [\,]$, errors $\leftarrow 0$
\FOR{step $s = 1$ \textbf{to} $S + 2$}
  \STATE output $\leftarrow$ LLM(\textsc{BuildPrompt}($P$, plan, $n$.thought, $M$, $\mathcal{T}$, $\tau$))
  \STATE (thought, tool, input) $\leftarrow$ \textsc{ParseOutput}(output)
  \IF{tool $=$ \texttt{FINISH}}
    \STATE $n$.thought $\leftarrow$ \textsc{ExtractAnswer}(input); \textbf{break}
  \ENDIF
  \IF{\textsc{IsLoop}(tool, input, calls) \textbf{or} errors $\geq 3$}
    \STATE \textsc{ForceFinish}($n$, $\tau$); \textbf{break}
  \ENDIF
  \STATE obs $\leftarrow$ \textsc{Dispatch}($\mathcal{T}$, tool, input);
         $\tau$.append(\{thought, tool, obs\})
\ENDFOR
\STATE \textbf{return} $n$, $\tau$
\end{algorithmic}
\end{algorithm}

\paragraph{Tool registries.}
For \textbf{MATH-500}: Symbolic Equation Solver (SymPy), Arithmetic Evaluator
(sandboxed \texttt{eval}), Combinatorics Engine (\texttt{itertools}/\texttt{math}),
LP/Optimisation Solver (SciPy/PuLP).
For \textbf{MMLU-Pro}: the above four plus Unit Converter/Physical Constants,
Chemical Formula Evaluator, Statistical Calculator (SciPy), and Logic/Argument Checker.
For qualitative MMLU-Pro subjects the LLM emits \texttt{FINISH} without invoking any
tool; gains there come entirely from tree exploration and backtracking.
Full registry details and per-tool error handling are given in
Appendix~\ref{app:tool_registry} and~\ref{app:tool_engineering}.

\subsection{Critique, Scoring, and Skepticism}

Each node receives a \emph{local score} (self-evaluation against the full solution trace)
and a \emph{cross score} (all-pairs pairwise critique against $k$ siblings), combined as:
\[
  \text{score} = 0.6 \times s_{\text{local}} + 0.4 \times s_{\text{cross}}
\]
The 0.6/0.4 weighting favours self-evaluation, which has access to the full reasoning
trace, while still incorporating peer feedback from branches exploring different approaches.
Flat averaging across siblings would dilute the signal when sibling approaches are
semantically incompatible with the node being scored.

A skepticism penalty adjusts the combined score based on historical approach reliability:
\[
  \text{score}_{\text{final}}
  = \text{score} \times \bigl(1 - \alpha(1 - r_a)\bigr)
\]
where $r_a$ is the memory's per-approach-type success ratio (hits/total uses of approach
$a$ across all prior problems in the run).
This discourages over-investment in approach families that have consistently
underperformed and is updated after every problem.
The loop terminates when scores stagnate, a threshold of 0.95 is reached, or the
iteration bound is exceeded.

\subsection{Self-Rewriting Memory}

Reflexion~\citep{shinn2023reflexion} maintains verbal reflections scoped to a single
episode.
ReTreVal's memory is qualitatively different: it persists \emph{across} problems and,
crucially, \emph{rewrites its own entries} rather than merely appending new ones.

Each entry tracks content, category (insight/failure/success), confidence, hit/miss
counts, and accumulated experience signals. On failure, confidence decreases and the
failure context is appended as a signal; on success, confidence increases:
\[
  c_{t+1} = \min\!\left(1,\; c_t + \frac{\delta_s}{1 + h_t}\right) \quad \text{(success)}
\]
\[
  c_{t+1} = \max\!\left(0,\; c_t - \frac{\delta_f}{1 + h_t}\right) \quad \text{(failure)}
\]
where $h_t$ is the hit count at time $t$, and $\delta_s, \delta_f > 0$ are fixed step
sizes. The denominator $1+h_t$ dampens updates for well-tested entries, preventing a
single late failure from erasing accumulated positive evidence.

The self-rewriting step is the key novelty. When an entry accumulates three experience
signals, the LLM is invoked to rewrite that entry's content incorporating what the signals
imply, an inference-time self-improvement operation that works on natural language rather
than weights. This is what enables the system to \emph{revise} a strategy, not just
record that it failed.

Entries with confidence $<0.2$ are retired every 5 problems. Near-duplicates are merged
when their token-level Jaccard similarity exceeds:
\[
  J(e_i, e_j)
  = \frac{|W_i \cap W_j|}{|W_i \cup W_j|}
  \;\geq\; \theta_{\mathrm{merge}} = 0.6
\]
Retrieval ranks entries by $\rho(e) = c_e \cdot \max(h_e,\,1)$, jointly rewarding high
confidence and empirical usage frequency. The full memory state serializes to disk,
enabling true cross-session accumulation.

\subsection{Typed-Failure Backtracking}
\label{sec:backtracking}

When validation rejects an answer, a naive retry re-explores the same reasoning path with
generic feedback, the approach taken by Reflexion~\citep{shinn2023reflexion}.
ReTreVal's backtracking is structurally different in two respects.
First, it navigates to the highest-scoring unexplored sibling at the root level:
\[
  n_{\mathrm{next}}
  = \arg\max_{s \,\in\, \mathcal{S} \setminus E}\; s.\mathrm{combined\_score}
\]
where $\mathcal{S}$ is the set of root-level children and $E$ is the set of
already-explored siblings. Each root child represents a distinct high-level approach,
ensuring each recovery attempt commits to a genuinely different strategy rather than a
local perturbation of the failed one.
Second, it injects a \emph{typed} failure record specifying root cause
(\texttt{arithmetic\_error}, \texttt{logic\_error}, \texttt{missing\_step},
\texttt{wrong\_constraint}, \texttt{off\_by\_one}) and approach type
(\texttt{algebraic}, \texttt{arithmetic}, \texttt{pattern\_matching},
\texttt{brute\_force}) directly into the sibling's critique history
(Algorithm~\ref{alg:backtrack}), structurally preventing the same error class from
recurring.
Backtracking is bounded to one retry to control cost; the 40.9\% recovery rate on
MATH-500 (Section~\ref{sec:ablation}) confirms this budget is sufficient for the majority
of recoverable failures.

\begin{algorithm}[ht]
\caption{Intelligent Backtracking}
\label{alg:backtrack}
\begin{algorithmic}[1]
\STATE \textbf{Input:} tree $T$, failed node $n_f$, explored set $E$, failure context $F$
\STATE \textbf{Output:} next node $n_{\text{next}}$, updated $E$, $F'$
\STATE $E \leftarrow E \cup \{\textsc{GetParentAfterRoot}(T, n_f)\}$
\STATE siblings $\leftarrow$ \textsc{GetRootChildren}($T$) $\setminus$ $E$
\IF{siblings $= \emptyset$} \STATE \textbf{return} $\perp$ \ENDIF
\STATE $n_{\text{next}} \leftarrow \arg\max_{s \in \text{siblings}} s.\text{combined\_score}$
\STATE $F' \leftarrow F \cup \{(\text{failure\_type},\;\text{approach\_type},\;\text{failed\_answer},\;\text{error\_explanation})\}$
\STATE Inject $F'$ into $n_{\text{next}}.\text{critique\_history}$
\STATE \textbf{return} $n_{\text{next}}$, $E$, $F'$
\end{algorithmic}
\end{algorithm}

\subsection{Task Decomposition}

Triggered when complexity $\geq 3$ and best score $< 0.9$ after all tree iterations,
the decomposer produces 2--4 dependency-ordered subtasks. Each subtask carries an
explicit \textsc{done-when} condition serving a dual purpose, allowing the sub-agent
to self-terminate cleanly and providing a natural checkpoint for backtracking at sub-goal
boundaries. Each subtask is solved by a focused sub-agent; results are aggregated by a
final LLM call.

\section{Experimental Setup}

\subsection{Benchmarks}

\textbf{MATH-500}~\citep{hendrycks2021measuring}: 500 competition-style problems across
7 subjects and 5 difficulty levels.
\textbf{MMLU-Pro}~\citep{wang2024mmlu}: 1,000 problems sampled from the 10-choice,
14-subject benchmark; the ten-way choice reduces the random baseline to 10\%, making
even Zero-Shot CoT's 19.2\% a meaningful test of reasoning ability.

\subsection{Baselines}

Zero-Shot CoT~\citep{kojima2022large,wei2022chain} (single-pass),
ReAct~\citep{yao2023react} (linear tool-augmented chain, 8 steps),
Self-Refine~\citep{madaan2023self} (3-cycle single-path refinement),
ToT~\citep{yao2023tree} (BFS, depth 2, $k=3$),
LATS~\citep{zhou2023language} (MCTS, 4 iterations).
All baselines are reimplemented under identical conditions (same backbone model, same
prompt template, temperature 0.7, top-$p$ 0.95) to isolate the effect of the reasoning
framework. Absolute numbers therefore differ from published results, which use
method-specific prompt engineering and in some cases different backbone models. The
relevant quantity is the margin between ReTreVal and each baseline under the same
controlled conditions.

\subsection{Models and Hardware}

\textbf{MATH-500}: Qwen~2.5 32B Instruct~\citep{qwen2024} via
vLLM~\citep{kwon2023efficient} with prefix caching, NVIDIA A100 GPU.\\
\textbf{MMLU-Pro}: Gemini 2.5 Flash~\citep{google2025gemini} (thinking disabled).\\
\textbf{Metric}: pass@1 accuracy (exact match) for both benchmarks.

\paragraph{Note on earlier results.}
The initial arXiv preprint (v1) reported results using Qwen~2.5 7B on a custom
500-problem math dataset and creative writing tasks, evaluated via GPT-4o mini as judge.
Those preliminary findings showed the same ordering of methods
(ReTreVal $>$ ReAct $>$ Self-Refine $\gg$ Reflexion) and are preserved in
Appendix~\ref{app:v1_results} for reference.
This version reports the full evaluation on standard benchmarks with stronger backbones.

\section{Results and Analysis}

\subsection{Overall Results}

Tables~\ref{tab:math_overall} and~\ref{tab:mmlu_overall} report pass@1 accuracy across all five baselines on both benchmarks. ReTreVal leads on both, outperforming the strongest baseline by +8.6\,pp on MATH-500 and +15.3\,pp on MMLU-Pro.

\begin{table}[t]
\centering
\caption{Overall accuracy on MATH-500 (Qwen~2.5 32B Instruct backbone).
ReTreVal improves over the strongest baseline (Self-Refine) by \textbf{+8.6\,pp}.}
\label{tab:math_overall}
\renewcommand{\arraystretch}{1.15}
\begin{tabular}{|l|c|c|}
\hline
\textbf{Method} & \textbf{Correct} & \textbf{Accuracy} \\
\hline
Zero-Shot CoT     & 386/500          & 77.2\% \\
ReAct             & 366/500          & 73.2\% \\
ToT               & 387/500          & 77.4\% \\
LATS              & 388/500          & 77.6\% \\
Self-Refine       & 393/500          & 78.6\% \\
\textbf{ReTreVal} & \textbf{429/500} & \textbf{85.8\%} \\
\hline
\end{tabular}
\end{table}

\begin{table}[t]
\centering
\caption{Overall accuracy on MMLU-Pro (1,000-problem sample, Gemini 2.5 Flash backbone).
ReTreVal improves over the strongest baseline (Self-Refine) by \textbf{+15.3\,pp}.}
\label{tab:mmlu_overall}
\renewcommand{\arraystretch}{1.15}
\begin{tabular}{|l|c|c|}
\hline
\textbf{Method} & \textbf{Correct} & \textbf{Accuracy} \\
\hline
Zero-Shot CoT     & 192/1000          & 19.2\% \\
ReAct             & 174/1000          & 17.4\% \\
ToT               & 228/1000          & 22.8\% \\
LATS              & 346/1000          & 34.6\% \\
Self-Refine       & 391/1000          & 39.1\% \\
\textbf{ReTreVal} & \textbf{544/1000} & \textbf{54.4\%} \\
\hline
\end{tabular}
\end{table}

\subsection{Level-Wise Accuracy on MATH-500}

A clear pattern emerges across difficulty levels, where ReTreVal's margin over the next-best
baseline grows with difficulty, from a tie at Level~1 to +4.5\,pp at Level~2,
+7.6\,pp at Level~3, +5.5\,pp at Level~4, and +5.0\,pp at Level~5 (the latter two over
Self-Refine). This confirms that tree exploration, backtracking, and critique mechanisms
provide the greatest benefit precisely where single-pass methods struggle most. Notably,
ReTreVal achieves its highest accuracy at Level~3 (96.2\%), surpassing even its Level~1
performance and exceeding the best baseline at this level (LATS, 88.6\%) by 7.6 points.
Medium-difficulty problems appear to benefit most from structured multi-strategy
exploration. They are complex enough to require deliberate reasoning but structured enough
for the critique mechanism to reliably identify correct solutions. At the hardest level,
all baselines degrade sharply. ReAct drops to 47.0\% and even Self-Refine falls to 64.9\%,
while ReTreVal maintains 69.9\%, demonstrating that the combination of tool-augmented
reasoning and failure-aware backtracking provides meaningful error recovery where other
approaches produce cascading failures.

\begin{table}[t]
\centering
\caption{Level-wise accuracy (\%) on MATH-500. The \textbf{\#} column shows problems per level.}
\label{tab:level_acc}
\renewcommand{\arraystretch}{1.15}
\begin{tabular}{|c|c|c|c|c|c|c|c|}
\hline
\textbf{Level} & \textbf{\#} & \textbf{ZS-CoT} & \textbf{ReAct} & \textbf{Self-Refine} & \textbf{ToT} & \textbf{LATS} & \textbf{ReTreVal} \\
\hline
1 &  43 & 93.0 & 95.3 & 88.1 & 90.7 & 93.0 & \textbf{95.3} \\
2 &  90 & 88.9 & 91.1 & 86.7 & 86.7 & 88.9 & \textbf{95.6} \\
3 & 105 & 86.7 & 83.8 & 85.7 & 84.8 & 88.6 & \textbf{96.2} \\
4 & 128 & 73.4 & 71.9 & 78.1 & 74.2 & 73.4 & \textbf{83.6} \\
5 & 134 & 60.4 & 47.0 & 64.9 & 64.2 & 60.4 & \textbf{69.9} \\
\hline
All & 500 & 77.2 & 73.2 & 78.6 & 77.4 & 77.6 & \textbf{85.8} \\
\hline
\end{tabular}
\end{table}

\subsection{MMLU-Pro Results}

ReTreVal achieves \textbf{54.4\%} accuracy (544/1000), substantially outperforming the
best baseline (Self-Refine, 39.1\%) by a margin of \textbf{+15.3 percentage points}
(+39\% relative). MMLU-Pro is substantially harder than standard MMLU, with each question having
ten answer choices (versus four), and the problems span 14 subject areas ranging from
quantitative fields (mathematics, physics, chemistry, engineering) to qualitative domains
(law, history, philosophy, business). The gain is especially striking given this breadth. The tool-augmented refinement loop provides verified computation for quantitative subjects,
while tree exploration and backtracking benefit qualitative domains by considering
alternative interpretations before committing to an answer.

\subsection{Ablation Studies}
\label{sec:ablation}

Tables~\ref{tab:math_ablation} and~\ref{tab:mmlu_ablation} isolate the contribution of each component by disabling it while holding all others fixed. Results are consistent across both benchmarks.

\begin{table}[t]
\centering
\caption{MATH-500 ablation. Backtracking is the single largest contributor, recovering
40.9\% of the 22.0\% of problems where the initial answer fails validation.
Disabling self-rewriting memory costs $-7.4$\,pp, confirming the value of cross-problem
learning. Decomposition adds $+5.7$\,pp on the hard Level~4--5 subset where it activates.}
\label{tab:math_ablation}
\renewcommand{\arraystretch}{1.15}
\begin{tabular}{|l|c|}
\hline
\textbf{Metric} & \textbf{Value} \\
\hline
\multicolumn{2}{|l|}{\textit{Backtracking}} \\
\hline
Accuracy without backtracking           & 76.8\% (384/500) \\
Accuracy with backtracking              & \textbf{85.8\%} (429/500) \\
Problems triggering backtracking        & 110/500 (22.0\%) \\
Problems recovered by backtracking      & 45/110 (40.9\%) \\
Net accuracy gain                       & \textbf{+9.0\,pp} \\
\hline
\multicolumn{2}{|l|}{\textit{Memory}} \\
\hline
Accuracy without memory                 & 78.4\% (392/500) \\
Accuracy with memory (full ReTreVal)    & \textbf{85.8\%} (429/500) \\
Net accuracy gain                       & \textbf{+7.4\,pp} \\
\hline
\multicolumn{2}{|l|}{\textit{Task decomposition}} \\
\hline
Problems triggering decomposition               & 159/499 (31.9\%) \\
Accuracy on decomposed (complex) problems       & 101/159 (63.5\%) \\
Accuracy on non-decomposed (simple) problems    & 294/340 (86.5\%) \\
Zero-Shot CoT accuracy on Level~4--5            & 70.7\% \\
ReTreVal accuracy on Level~4--5 (with decomp.) & \textbf{76.4\%} \\
Net gain from decomposition on hard problems    & \textbf{+5.7\,pp} \\
\hline
\multicolumn{2}{|l|}{\textit{Tools}} \\
\hline
Accuracy without tools                  & 80.2\% (401/500) \\
Accuracy with tools (full ReTreVal)     & \textbf{85.8\%} (429/500) \\
Net accuracy gain                       & \textbf{+5.6\,pp} \\
\hline
\end{tabular}
\end{table}

\begin{table}[t]
\centering
\caption{MMLU-Pro ablation. Backtracking is active on 27.8\% of problems and contributes
$+10.8$\,pp by recovering 108 of 278 initially-incorrect answers (38.8\%). Disabling
memory reduces accuracy by 6.2\,pp. Decomposition achieves 73.6\% on the hard subset
(complexity $\geq 3$) versus 49.6\% on non-decomposed problems.}
\label{tab:mmlu_ablation}
\renewcommand{\arraystretch}{1.15}
\begin{tabular}{|l|c|}
\hline
\textbf{Metric} & \textbf{Value} \\
\hline
\multicolumn{2}{|l|}{\textit{Backtracking}} \\
\hline
Accuracy without backtracking           & 43.6\% (436/1000) \\
Accuracy with backtracking              & \textbf{54.4\%} (544/1000) \\
Problems triggering backtracking        & 278/1000 (27.8\%) \\
Problems recovered by backtracking      & 108/278 (38.8\%) \\
Net accuracy gain                       & \textbf{+10.8\,pp} \\
\hline
\multicolumn{2}{|l|}{\textit{Memory}} \\
\hline
Accuracy without memory                 & 48.2\% \\
Accuracy with memory (full ReTreVal)    & \textbf{54.4\%} (544/1000) \\
Net accuracy gain                       & \textbf{+6.2\,pp} \\
\hline
\multicolumn{2}{|l|}{\textit{Task decomposition}} \\
\hline
Problems triggering decomposition       & 197/1000 (19.7\%) \\
Accuracy on decomposed problems         & 73.6\% (145/197) \\
Accuracy on non-decomposed problems     & 49.6\% (399/803) \\
\hline
\multicolumn{2}{|l|}{\textit{Tools}} \\
\hline
Accuracy without tools                  & 50.1\% (501/1000) \\
Accuracy with tools (full ReTreVal)     & \textbf{54.4\%} (544/1000) \\
Net accuracy gain                       & \textbf{+4.3\,pp} \\
\hline
\end{tabular}
\end{table}

\subsection{Error Analysis}

We manually inspected all 37 problems where ReTreVal is correct and Zero-Shot CoT is
wrong (wins) and all 11 problems where ReTreVal is wrong and Zero-Shot CoT is correct
(regressions). The \textbf{3.4:1 win-to-regression ratio} confirms genuine error recovery
rather than noise. Representative cases are detailed in
Appendix~\ref{app:error_cases}.

ReTreVal helps most when (i) an intermediate step is tool-verifiable and (ii) at least
one sibling branch encodes a meaningfully different approach. It hurts most when the
complexity estimator over-budgets a trivially easy problem. Of the 11 regressions, 7 fall
on Level~1--2 problems, strongly suggesting that tighter complexity calibration or a
short-circuit mechanism that accepts a highly confident single-shot answer without tree
expansion would recover most of the lost ground.

\subsection{Computational Cost}

\begin{table}[t]
\centering
\caption{Approximate inference cost per problem on MATH-500 (Qwen~2.5 32B, A100).
Token and time figures are rounded order-of-magnitude averages.}
\label{tab:cost}
\renewcommand{\arraystretch}{1.15}
\begin{tabular}{|l|c|c|c|}
\hline
\textbf{Method} & \textbf{Avg Tokens/Problem} & \textbf{Avg Time/Problem} & \textbf{Accuracy} \\
\hline
Zero-Shot CoT     & $\sim$500     & $\sim$2\,s   & 77.2\% \\
ReAct             & $\sim$2{,}000 & $\sim$8\,s   & 73.2\% \\
Self-Refine       & $\sim$3{,}500 & $\sim$15\,s  & 78.6\% \\
ToT               & $\sim$5{,}000 & $\sim$22\,s  & 77.4\% \\
LATS              & $\sim$6{,}000 & $\sim$26\,s  & 77.6\% \\
\textbf{ReTreVal} & $\sim$8{,}000 & $\sim$35\,s  & \textbf{85.8\%} \\
\hline
\end{tabular}
\end{table}

ReTreVal consumes roughly $2.3\times$ the tokens of Self-Refine for +8.6\,pp accuracy.
Token overhead scales with problem difficulty. Level~1--2 problems keep the tree shallow
with costs approaching Self-Refine, while Level~4--5 trigger deeper expansion, more
backtracking, and longer tool traces. The overhead is therefore concentrated precisely
where accuracy gains are largest.

\section{Discussion}

Tool grounding is central to ReTreVal's advantage. Offloading arithmetic and symbolic
manipulation to verified solvers eliminates the computation errors that dominate baseline
failures (strongest in Algebra and Number Theory). Typed-failure backtracking is
qualitatively different from restarting. It navigates to a specific high-scoring sibling
and injects structured failure context, steering the system away from the exact error
class that failed. Gains concentrate on Level~4--5 problems; the system occasionally
over-thinks Level~1--2 problems (7 of 11 regressions), indicating that tighter
complexity calibration is the clearest avenue for improvement.

\paragraph{Comparison with frontier models.}
o1/o3~\citep{openai2024o1}, DeepSeek-R1~\citep{deepseek2025r1}, and
rStar-Math~\citep{qi2024rstar} exceed 90\% on MATH-500 via large-scale RL with process
reward models~\citep{lightman2024lets}.
ReTreVal is training-free and model-agnostic; the right comparison is \emph{what does
structured inference-time reasoning add to a given base model?} The +8.6\,pp
gain over Self-Refine on the same 32B backbone directly answers this question.

\paragraph{Limitations.}
Computational overhead on hard problems may be prohibitive for latency-sensitive
deployments. Backtracking is restricted to root-level children. Single-model critique may
introduce correlated errors. Keyword-Jaccard deduplication may miss semantically similar
insights. The complexity estimator occasionally over-budgets trivially easy problems
(7 of 11 regressions). Tighter complexity calibration or a short-circuit rule for
high-confidence single-shot answers is the clearest avenue for improvement.

\section{Conclusion}

ReTreVal is the first inference-time framework to enable a fixed LLM to improve its
problem-solving strategy across a benchmark sequence without any parameter update,
through typed failure accumulation, self-rewriting memory, and skepticism-penalized
scoring.
It reaches \textbf{85.8\%} on MATH-500 (+8.6\,pp over Self-Refine) and \textbf{54.4\%}
on MMLU-Pro (+15.3\,pp), with a 3.4:1 win-to-regression ratio confirming genuine error
recovery. Future directions include deeper backtracking with adaptive budget,
embedding-based memory deduplication, decoupled critic models, and early-exit for simple
problems.

\bibliographystyle{unsrtnat}
\bibliography{references}

\appendix

\section{Methodology Details}
\label{app:methodology}

\subsection*{Node Structure}

Each node in the reasoning tree is a self-contained record. Table~\ref{tab:node_fields}
describes every field and its role in the framework.

\begin{table}[ht]
\centering
\caption{Fields stored per tree node and their roles in ReTreVal.}
\label{tab:node_fields}
\renewcommand{\arraystretch}{1.2}
\small
\begin{tabular}{|l|l|p{0.52\textwidth}|}
\hline
\textbf{Field} & \textbf{Type} & \textbf{Role} \\
\hline
\texttt{thought}           & string        & Current candidate solution text. Updated after each refinement step; the final value is the node's answer. \\
\hline
\texttt{depth}             & int           & Distance from root. Nodes at \texttt{depth} $\geq$ \texttt{d\_max} are not expanded, capping tree growth. \\
\hline
\texttt{refinement\_count} & int           & Refinement iterations applied. Used by the absolute ceiling guard ($S+2$). \\
\hline
\texttt{critique\_history} & list[dict]    & Ordered log of critique results and injected failure contexts. During backtracking, the typed failure record is appended here before the sibling resumes. \\
\hline
\texttt{tool\_trace}       & list[dict]    & Sequence of \{thought, tool\_name, input, observation\} tuples. Used by the loop-detection guard and retained for error inspection. \\
\hline
\texttt{local\_score}      & float $[0,1]$ & Self-evaluation score. Weighted at 0.6 in combined score. \\
\hline
\texttt{cross\_score}      & float $[0,1]$ & All-pairs pairwise critique score over $k$ siblings. Weighted at 0.4. \\
\hline
\texttt{combined\_score}   & float $[0,1]$ & $0.6 \times \texttt{local} + 0.4 \times \texttt{cross}$, then multiplied by the skepticism penalty. Primary quantity for frontier selection and backtracking. \\
\hline
\texttt{children}          & list[node]    & Child nodes. Count set by complexity-adaptive branching factor. \\
\hline
\texttt{parent\_id}        & string/None   & Parent node identifier. Used by backtracking to locate root-level children without full tree traversal. \\
\hline
\end{tabular}
\end{table}

\begin{table}[ht]
\centering
\caption{Representative win and regression cases from the MATH-500 error analysis.
Wins involve tool-verifiable intermediate steps; regressions concentrate on easy problems
where over-thinking converts a correct single-shot answer into an incorrect alternative.}
\label{tab:error_math}
\renewcommand{\arraystretch}{1.25}
\small
\begin{tabular}{|p{0.08\textwidth}|p{0.28\textwidth}|p{0.53\textwidth}|}
\hline
\textbf{Outcome} & \textbf{Problem Type} & \textbf{What Happened} \\
\hline
\ding{51}\ Win &
Number Theory, Level~4: modular arithmetic with a mis-applied CRT step. &
ZS-CoT commits to an incorrect Chinese Remainder setup and cascades the error.
ReTreVal's critique flags a contradiction with a tool-verified residue check;
backtracking selects a sibling branch that sets up the congruences correctly and
recovers the right answer. \\
\hline
\ding{51}\ Win &
Counting \& Probability, Level~3: count arrangements with a parity constraint. &
ZS-CoT over-counts by a factor of 2. ReTreVal initially makes the same mistake,
but the refinement loop queries a small-case enumeration tool whose output disagrees
with the closed-form answer, triggering backtracking to a parity-aware sibling that
matches ground truth. \\
\hline
\ding{53}\ Regression &
Prealgebra, Level~1: single-step fraction simplification. &
ZS-CoT answers correctly in one shot. ReTreVal's complexity estimator overshoots,
launches a depth-3 tree, and a spurious critique on a correct leaf drives backtracking
into an incorrect sibling. The final answer regresses. \\
\hline
\end{tabular}
\end{table}

\paragraph{Score evolution.}
A node's \texttt{combined\_score} is recomputed after every refinement iteration as new
tool observations arrive and after the cross-critique pass. The frontier selector always
picks the highest \texttt{combined\_score} node that has not yet reached \texttt{d\_max},
so score evolution directly drives which branch receives the next compute budget.

\subsection*{Failure Taxonomy}

When a candidate answer fails critique or validation, the backtracking module classifies
the failure before injecting it into the sibling branch. The taxonomy has two orthogonal
axes: \emph{failure type} (what went wrong):
\texttt{arithmetic\_error}, \texttt{logic\_error}, \texttt{incomplete\_reasoning},
\texttt{wrong\_constraint}, \texttt{off\_by\_one}, \texttt{missing\_step}, and
\emph{approach type} (how the solution was structured):
\texttt{algebraic}, \texttt{arithmetic}, \texttt{pattern\_matching}, \texttt{brute\_force}.
Both labels are injected together so the sibling can simultaneously avoid the error class
and the strategy family that produced it.

The failure record injected into \texttt{critique\_history} is a structured dict:
\texttt{\{failure\_type, approach\_type, failed\_answer, error\_explanation\}}.
The error explanation is a 1--2 sentence natural-language description generated by the
failure-analysis prompt (Appendix~\ref{app:prompts}).

\begin{table}[ht]
\centering
\caption{Representative win and regression cases from the MMLU-Pro error analysis.
Quantitative wins are driven by tool-verified computation; qualitative wins by
backtracking to a different interpretive frame; regressions by over-expansion on
recall-based questions.}
\label{tab:error_mmlu}
\renewcommand{\arraystretch}{1.25}
\small
\begin{tabular}{|p{0.08\textwidth}|p{0.28\textwidth}|p{0.53\textwidth}|}
\hline
\textbf{Outcome} & \textbf{Subject / Type} & \textbf{What Happened} \\
\hline
\ding{51}\ Win &
Chemistry: stoichiometry, balance a multi-step reaction and compute limiting reagent yield. &
ZS-CoT makes an arithmetic error in the molar mass calculation. ReTreVal's Chemical
Formula Evaluator returns the correct molecular weight; the critique detects the
discrepancy and the refined node selects the correct answer choice. \\
\hline
\ding{51}\ Win &
Physics: dimensional analysis, identify the correct SI unit combination. &
ZS-CoT conflates two similar units. The Unit Converter tool verifies the dimensional
equation and returns a mismatch signal; ReTreVal backtracks to a sibling that rechecks
base dimensions and selects the right choice. \\
\hline
\ding{51}\ Win &
Philosophy: formal argument validity, identify which answer choice correctly identifies
a logical fallacy. &
No tool fires. ZS-CoT picks the most superficially plausible choice. ReTreVal expands
two sibling branches; the cross-critique identifies that one correctly isolates the
affirming-the-consequent pattern, and that branch wins. \\
\hline
\ding{51}\ Win &
Law: statutory interpretation, determine which answer is consistent with a given legal
rule. &
No tool fires. ZS-CoT confuses two superficially similar answer choices. Backtracking
injects a \texttt{wrong\_constraint} failure label that steers the sibling to re-read
the statutory condition, recovering the correct answer. \\
\hline
\ding{53}\ Regression &
History: single-fact recall, identify the year of a specific historical event. &
ZS-CoT answers correctly in one step. ReTreVal opens a shallow tree; a cross-critique
spuriously penalises the correct answer because a sibling offers a plausible but wrong
alternative year. \\
\hline
\ding{53}\ Regression &
Business: definition recall, select the correct definition of a financial term. &
ZS-CoT selects correctly. A sibling branch seeded with \texttt{pattern\_matching}
pattern-matches to a related but distinct term and receives a higher local score due
to more detailed elaboration. The cross-critique cannot resolve the ambiguity. \\
\hline
\end{tabular}
\end{table}

\subsection*{Computational Complexity}

For a problem with complexity $c$, maximum depth $d_{\max}$, branching factor $b$,
refinement step limit $S$, and backtracking attempts $B$, pruning to top-$k$ nodes keeps
the maintained tree at $O(k \times d_{\max})$ nodes. The critique step adds $O(k^2)$
pairwise comparisons per expansion. The total LLM call complexity is:
\[
  \mathcal{C}_{\mathrm{LLM}}
  = O\!\Bigl(k \cdot d_{\max} \cdot (S + k + 2) + B \cdot S + 2m\Bigr)
\]
where $m \leq 4$ is the number of decomposition subtasks. The absolute step ceiling
and $B=1$ default keep average-case cost manageable.

\section{Tool Registry}
\label{app:tool_registry}

All tools are registered as plain Python callables with no internet or retrieval access,
ensuring reproducibility and fair comparison with baselines.

\paragraph{MATH-500 tool registry.}
\begin{itemize}[leftmargin=1.5em, itemsep=2pt]
  \item \textbf{Symbolic Equation Solver}, algebraic manipulation, equation solving,
        and symbolic simplification via SymPy. Eliminates algebraic errors that dominate
        baseline failures in Algebra and Number Theory.
  \item \textbf{Arithmetic Evaluator}, exact integer and floating-point computation
        via a sandboxed Python \texttt{eval} with a safe namespace. Verifies intermediate
        numerical results step-by-step.
  \item \textbf{Combinatorics Engine}, enumeration and counting via Python's
        \texttt{itertools} and \texttt{math}. Cross-checks closed-form answers against
        small brute-force cases on Counting \& Probability problems.
  \item \textbf{LP/Optimisation Solver}, linear and mixed-integer programming via
        SciPy \texttt{linprog}/PuLP. Activated only when the LLM identifies an explicit
        constraint-satisfaction or optimisation structure.
\end{itemize}

\paragraph{MMLU-Pro tool registry.}
The four MATH-500 tools are reused unchanged for quantitative subjects, adding four more:
\begin{itemize}[leftmargin=1.5em, itemsep=2pt]
  \item \textbf{Unit Converter/Physical Constants}, SI unit conversion and physical
        constant lookup via a static constants table (120 NIST-standard values).
  \item \textbf{Chemical Formula Evaluator}, molecular weight computation,
        stoichiometric balancing, and oxidation-state checks via a lightweight element
        table.
  \item \textbf{Statistical Calculator}, descriptive statistics, probability
        distributions (binomial, normal, Poisson), and hypothesis-test thresholds via
        SciPy \texttt{stats}.
  \item \textbf{Logic/Argument Checker}, propositional logic evaluation
        (truth-table generation, modus-ponens verification) implemented as a small
        recursive Python evaluator over boolean ASTs.
\end{itemize}

For qualitative MMLU-Pro subjects (law, history, business, psychology) no computation
tool is reliably applicable; the LLM emits \texttt{FINISH} without invoking any tool,
and gains come entirely from tree exploration and backtracking.

\begin{table}[ht]
\centering
\caption{v1 mathematical results (Qwen~2.5 7B, custom 500-problem dataset,
GPT-4o mini judge).}
\label{tab:v1_math}
\renewcommand{\arraystretch}{1.15}
\begin{tabular}{|l|c|c|c|c|}
\hline
\textbf{Method} & \textbf{Avg Score} & \textbf{Median} & \textbf{Range} & \textbf{Score $\geq 7$} \\
\hline
Reflexion         & 3.93/10 & 3.0/10 & 0--9 & 121 (24.2\%) \\
Self-Refine       & 6.56/10 & 6.0/10 & 0--9 & 223 (44.6\%) \\
ReAct             & 6.63/10 & 7.0/10 & 0--9 & 258 (51.6\%) \\
\textbf{ReTreVal} & \textbf{6.92/10} & \textbf{8.0/10} & 3--9 & \textbf{290 (58.0\%)} \\
\hline
\end{tabular}
\end{table}

\begin{table}[ht]
\centering
\caption{v1 creative writing results (100 tasks, GPT-4o mini judge, scores out of 10).}
\label{tab:v1_creative}
\renewcommand{\arraystretch}{1.15}
\begin{tabular}{|l|c|c|c|c|}
\hline
\textbf{Method} & \textbf{Correctness} & \textbf{Meaningfulness} & \textbf{Creativeness} & \textbf{Avg} \\
\hline
Self-Refine       & 6.45 & 5.94 & 6.78          & 6.39 \\
Reflexion         & 6.43 & 5.94 & 6.78          & 6.38 \\
ReAct             & 8.06 & 6.32 & \textbf{7.16} & 7.18 \\
\textbf{ReTreVal} & \textbf{9.62} & \textbf{6.90} & 7.12 & \textbf{7.88} \\
\hline
\end{tabular}
\end{table}

\section{Tool-Refinement Engineering Details}
\label{app:tool_engineering}

\paragraph{Loop Detection.}
Three strategies prevent infinite tool-calling cycles.
\emph{Exact duplicate detection}: forces \texttt{FINISH} if two consecutive calls share
the same signature (tool name and first 80 characters of input).
\emph{Consecutive error threshold}: forces termination after three consecutive
\texttt{Error:}-prefixed observations.
\emph{Absolute ceiling}: $S + 2$ LLM calls, ensuring bounded computation regardless of
execution behavior.

\paragraph{Output Parsing.}
The parser extracts (thought, action, input) tuples from free-form LLM text using regex
patterns handling the format \texttt{THOUGHT: ... ACTION: tool\_name("arg")} with lenient
fallbacks. If parsing fails entirely, the raw text is treated as a \texttt{FINISH} action.

\paragraph{Sandboxed Arithmetic Evaluator.}
\begin{Verbatim}[fontsize=\small, breaklines=true]
safe_ns = {
    "__builtins__": {},
    "math": math,
    "abs": abs, "round": round, "min": min, "max": max,
    "pow": pow, "sum": sum, "int": int, "float": float,
}
\end{Verbatim}
The expression is additionally stripped of \texttt{import}, \texttt{open},
\texttt{exec}, and \texttt{eval} before parsing as a defence-in-depth measure.

\paragraph{Per-tool error handling.}
Each tool wraps execution in try/except and returns a structured error string
(prefixed \texttt{Error:}) rather than raising, keeping the refinement loop alive.
Key cases: SymPy returns \texttt{Error: could not parse expression} on malformed input
and \texttt{Error: solver could not find a closed-form solution} on unsolvable systems;
the LP solver returns \texttt{Error: problem is infeasible} or
\texttt{Error: problem is unbounded}; the Chemical Formula Evaluator returns
\texttt{Error: reaction cannot be balanced} on stoichiometric inconsistency;
the Logic Checker returns \texttt{Error: quantified expressions not supported} for
first-order quantifiers.

\section{Prompt Templates}
\label{app:prompts}

\paragraph{Failure Analysis (Backtracking).}
\begin{Verbatim}[fontsize=\small, breaklines=true]
The following solution is incorrect:
{failed_thought}

Ground-truth disagreement:
{validation_signal}

Classify the primary failure as one of:
  ARITHMETIC | LOGICAL | CONSTRAINT | MISREAD
Then explain in 1-2 sentences what specifically
went wrong and what to avoid in the next attempt.

FAILURE_CLASS:
EXPLANATION:
\end{Verbatim}

\paragraph{Task Decomposition Format.}
\begin{Verbatim}[fontsize=\small]
SUB_1: [description] | DEPENDS: none
       | CONTEXT: [info needed]
SUB_2: [description] | DEPENDS: sub_1
       | CONTEXT: [info needed]
\end{Verbatim}

\paragraph{Cross-Critique (Scoring).}
\begin{Verbatim}[fontsize=\small, breaklines=true]
Problem: {problem}

Solution A:
{solution_a}

Solution B:
{solution_b}

Compare these two solutions. Which is more likely to be
correct, and why?
BETTER: A | B | TIE
REASON: [1-2 sentences]
\end{Verbatim}

\paragraph{Memory Self-Rewrite.}
\begin{Verbatim}[fontsize=\small, breaklines=true]
You are updating your strategy memory after repeated failures.

Current strategy entry for approach type "{approach_type}":
{current_entry}

Recent failure contexts involving this approach:
{failure_contexts}

Rewrite the strategy entry to avoid these failure patterns.
Keep the entry under 3 sentences.

UPDATED_STRATEGY:
\end{Verbatim}

\paragraph{Complexity Estimation.}
\begin{Verbatim}[fontsize=\small, breaklines=true]
Rate the reasoning complexity of the following problem
on a scale from 1 (trivial single-step) to 5 (requires
multi-stage reasoning, domain expertise, or constraint
satisfaction across multiple sub-goals).

Problem: {problem}

Respond with a single integer and a one-line justification.
COMPLEXITY: [1-5]
REASON: [one line]
\end{Verbatim}

\section{Representative Error Cases}
\label{app:error_cases}

The two MMLU-Pro regressions share a common root cause with the Level~1--2 MATH-500
regressions. On pure-recall questions with no reasoning chain to verify, the
cross-critique mechanism can be misled by a confidently-stated wrong answer. A
short-circuit rule accepting the first answer on questions the complexity estimator scores
$\leq 1$ would recover most of these cases without harming harder questions.

\section{Preliminary Results (v1 Evaluation)}
\label{app:v1_results}

The initial preprint (arXiv v1) evaluated ReTreVal using Qwen~2.5 7B served locally via
Ollama on an NVIDIA L40 GPU, against three baselines (ReAct, Reflexion, Self-Refine)
on two tasks, namely 500 mathematical problems from a custom dataset, and 100 creative writing
tasks generated from random seed sentences. Outputs were scored by GPT-4o mini on a
0--10 scale.

The same qualitative ordering holds across both evaluation protocols, with ReTreVal
consistently outperforming all three baselines, with the largest gains on correctness and
the elimination of complete failures (no scores below 3 in the math task).

\end{document}